\documentclass[letterpaper,10pt,journal,twoside]{IEEEtran}

\usepackage{graphicx}
\usepackage{bm}
\usepackage{cite}
\input{preamble}

\title{Extending Wire-Driven Parallel Robots to Multi-Floating-Link Systems with an Inter-Body~Wire}

\author{%
Shintaro Inoue$^{1}$, Kento Kawaharazuka$^{1}$, Temma Suzuki$^{1}$, Sota Yuzaki$^{1}$, Keita Yoneda$^{1}$, and Kei Okada$^{1}$%
\thanks{%
Manuscript received: April 14, 2026; Revised: June 17, 2026; Accepted: July 27, 2026.%
}%
\thanks{%
This paper was recommended for publication by Editor M. Ani Hsieh upon evaluation of the Associate Editor and Reviewers' comments.%
}%
\thanks{%
$^{1}$The authors are with the Department of Mechano-Informatics, Graduate School of Information Science and Technology, The University of Tokyo, 7-3-1 Hongo, Bunkyo-ku, Tokyo 113-8656, Japan {\tt\footnotesize [s-inoue, kawaharazuka, t-suzuki, yuzaki, yoneda, k-okada]@jsk.imi.i.u-tokyo.ac.jp}.%
}%
\thanks{Digital Object Identifier (DOI): see top of this page.}%
}
\usepackage{eso-pic}
\AddToShipoutPictureFG*{\AtPageLowerLeft{\put(36,12){\parbox[b]{\dimexpr\paperwidth-72pt\relax}{\fontsize{6}{7}\selectfont\raggedright\hyphenpenalty=10000\exhyphenpenalty=10000 \textcopyright{} 2026 IEEE. Personal use of this material is permitted. Permission from IEEE must be obtained for all other uses, in any current or future media, including reprinting/republishing this material for advertising or promotional purposes, creating new collective works, for resale or redistribution to servers or lists, or reuse of any copyrighted component of this work in other works. S. Inoue, K. Kawaharazuka, T. Suzuki, S. Yuzaki, K. Yoneda and K. Okada, "Extending Wire-Driven Parallel Robots to Multi-Floating-Link Systems With an Inter-Body Wire," in IEEE Robotics and Automation Letters, vol. 11, no. 10, pp. 11314-11321, Oct. 2026, doi: 10.1109/LRA.2026.3723331. DOI: https://doi.org/10.1109/LRA.2026.3723331}}}}
\AddToShipoutPictureFG*{\AtPageUpperLeft{\put(36,-43){\fontsize{6}{7}\selectfont DOI: https://doi.org/10.1109/LRA.2026.3723331}}}
\begin{document}

\maketitle

\begin{abstract}
    Parallel wire-driven systems are a framework in which multiple wires are connected to an end effector,
    and the end effector is driven by the tensions of those wires.
    Because they can transmit large forces, up to the breaking strength of the wires,
    over lengths on the order of several tens of meters,
    they can realize powerful actuation in large workspaces.
    In recent years, studies have also investigated connecting wires to external environments
    such as trees and columns and forming parallel wire actuation on site.
    However, conventional parallel wire-driven systems often assume a single end effector,
    and because of interference between wires and links in a shared workspace,
    it has been difficult to handle multiple floating links simultaneously.
    In this study, we realize multi-floating-link parallel wire actuation by introducing an inter-body wire.
    In real-robot experiments,
    the two bodies supported and lifted a mannequin and transferred it onto a stretcher as a rescue motion.
    This shows that parallel wire actuation can be extended from a single end effector to a multi-floating-link system,
    and indicates its potential for broader tasks such as the manipulation of large objects and rescue support.
\end{abstract}


\section{Introduction}\label{sec:introduction}
  \IEEEPARstart{C}{able-Driven} Parallel Robots (CDPRs) are robots that control the position and orientation of an end effector
  by connecting wires extending from multiple winches to the end effector
  and adjusting the length and tension of each wire.
  Many CDPRs have been developed so far.
  For example, CoGiRo\cite{7050380} connects a box-shaped end effector to a frame with multiple wires
  and moves it in three-dimensional space by winding those wires \cite{10.1007/978-3-319-09489-2_6, 7989723}.
  By mounting a robot arm on the end effector,
  it realizes operation in a large three-dimensional space in a warehouse\cite{8967836}.
  In addition, Pylos\cite{IAAC_Pylos}, which uses a 3D printer head as the end effector,
  realizes a 3D printer with a large workspace of $13.6 \times 9.4 \times 3.3$ m\cite{Izard2017}.
  Many of these conventional CDPRs adopt a configuration
  in which wires are wound by winches installed on a fixed frame or on an indoor structure.
  In recent years, environment-anchored CDPRs\cite{inoue2024cubix,cubix} have also been proposed,
  in which the robot itself, as the end effector, carries the winches
  and drives itself by connecting wires extending from its body to the external environment,
  such as trees and columns.

  \begin{figure}[t]
    \begin{center}
      \includegraphics[width=0.6\columnwidth]{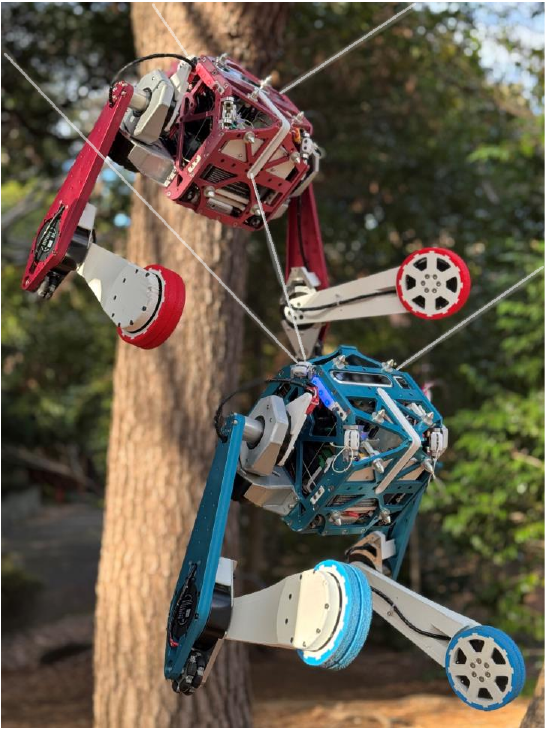}
      \vspace{-1.0ex}
      \caption{
        Overview of the TwinX, a multi-floating-link parallel wire-driven system.
        Two wheeled-legged robots are connected by an inter-body wire, and each robot can also
        anchor wires to the environment.
      }
      \vspace{-5ex}
      \label{fig:fig1}
    \end{center}
  \end{figure}

  Many of these conventional CDPRs
  have focused on configurations in which a single floating-link end effector is driven by multiple wires.
  In contrast, to support, transport, or manipulate larger objects,
  a configuration that supports the object from multiple points may become important.
  In fact, as shown by examples in which multiple humanoid robots cooperatively transport an object\cite{6907355,Wu17012016},
  and by examples in which multiple bipedal robots transport a common payload\cite{pandit25a},
  multiple robots can cooperate to handle large objects that are difficult to handle with a single robot.

  Even in CDPRs, if multiple floating links could be driven cooperatively,
  they would be effective for the manipulation and transport of objects that require multipoint support.
  However, CDPRs require multiple wires for actuation,
  and because those wires span a large region of the workspace,
  it is not easy to handle multiple floating links in the same workspace.
  In particular, as the number of floating links increases,
  wire interference and the difficulty of wire arrangement design also increase.
  As studies related to multi-robot CDPRs,
  there are two robot arms moving on a shared cable\cite{11127592}
  and a dual-arm robot arm based on an equivalent mechanism\cite{Cheng2025},
  but their motions are limited to a straight line,
  and the wide workspace and high degree of freedom inherent to CDPRs are not fully utilized.

  Therefore, in this study, we propose TwinX, shown in \figref{fig:fig1},
  as a multi-floating-link parallel wire-driven system with an inter-body wire.
  TwinX is a system in which two wheeled-legged robots are connected by a single inter-body wire,
  and each robot can connect its wires to the external environment.
  As a whole, the system is configured as an environment-anchored CDPR
  in which the two robots serve as floating links.
  This configuration forms a system in which two floating links
  cooperatively operate in a shared workspace by parallel wire actuation.
  In real-robot experiments, we performed a task that mimics rescue activity,
  in which a mannequin is loaded onto a stretcher and transported,
  and realized a transport task with multipoint support using the proposed system.

  The contribution of this study is that
  it extends CDPRs, which have conventionally driven a single floating link,
  to a framework in which multiple floating links cooperatively operate, and shows this on a real robot.
  This shows a first step toward extending the advantages of CDPRs
  to the transport and manipulation of objects that require multipoint support.

\section{Design Concept and Hardware Configuration} \label{sec:hardware}
\subsection{Effect of the Inter-Body Wire on the Workspace}
  \begin{figure}[t]
    \begin{center}
      \includegraphics[width=0.75\columnwidth]{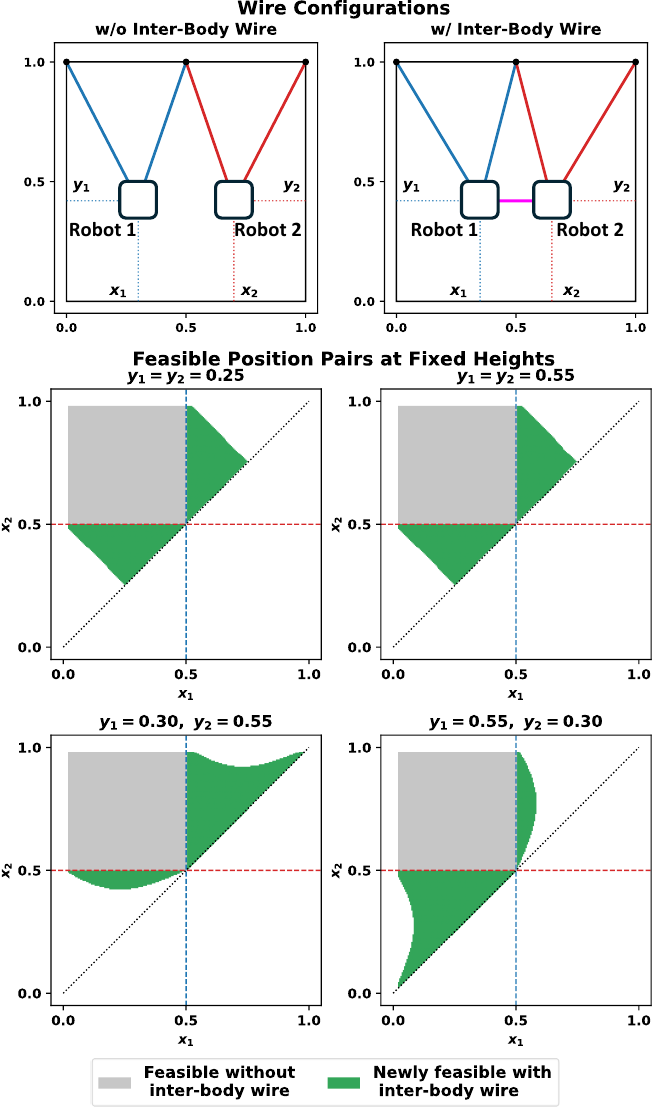}
      \vspace{-1.0ex}
      \caption{
        Representative example illustrating the effect of the inter-body wire on feasible workspace.
        The top row shows wire configurations without and with the inter-body wire.
        The bottom row shows feasible pairs of horizontal positions $(x_1, x_2)$ for fixed heights $(y_1, y_2)$.
        Gray indicates pairs that are feasible without the inter-body wire,
        while green indicates pairs that become newly feasible after adding the inter-body wire.
      }
      \vspace{-4.0ex}
      \label{fig:poc}
    \end{center}
  \end{figure}

  To confirm the effect of the inter-body wire on the workspace,
  we discuss a representative configuration example.
  Specifically, in a two-dimensional plane, the two bodies are represented as point masses,
  and the baseline configuration is that each body is connected to two upper points in the environment.
  We then consider the case in which,
  instead of introducing a new environment-anchoring wire,
  an inter-body wire is introduced between the two bodies.

  In the simplified model in this subsection,
  each wire is assumed to be an ideal wire that transmits only tension.
  In addition, a body configuration is defined as feasible
  if there exists a set of wire tensions that satisfies static equilibrium against gravity at that configuration.

  The top row of \figref{fig:poc} shows the wire arrangements in this representative example without and with the inter-body wire.
  In the bottom row of \figref{fig:poc}, for multiple fixed-height conditions $(y_1, y_2)$,
  we visualize whether the horizontal position pair $(x_1, x_2)$ of the two bodies is feasible in the above sense.
  The gray region represents body-position pairs that are feasible even without the inter-body wire,
  and the green region represents body-position pairs that become newly feasible by introducing the inter-body wire.
  This shows that the inter-body wire expands the feasible body-position pairs,
  and this tendency is commonly observed under multiple symmetric and asymmetric fixed-height conditions.

  The important point here is that this effect is not simply due to ``adding one more wire,''
  but rather to the fact that ``the additional connection is between the bodies.''
  To obtain a similar tension direction with an additional environment-anchoring wire,
  it would be necessary to prepare a new anchor point in the external environment,
  and that wire would also tend to cross a region occupied by the bodies or by other wires.
  In contrast, the inter-body wire provides an additional tension direction by using the other body,
  which is already part of the system, as the connection target,
  and therefore realizes a multi-floating-link parallel wire-driven system without interfering with the bodies or other wires.

  Although the above example is based on a simplified two-dimensional point-mass model,
  the same idea can be interpreted for the actual three-dimensional system
  by considering the 6D wrench equilibrium acting on each body.
  However, the actual usable workspace is smaller than
  that of the idealized model because of physical constraints in the real system,
  such as the orientation limits of each platform and the lower and upper bounds of the wire tensions.

\subsection{Overall Hardware Configuration}
  \begin{figure}[t]
    \begin{center}
      \includegraphics[width=1.0\columnwidth]{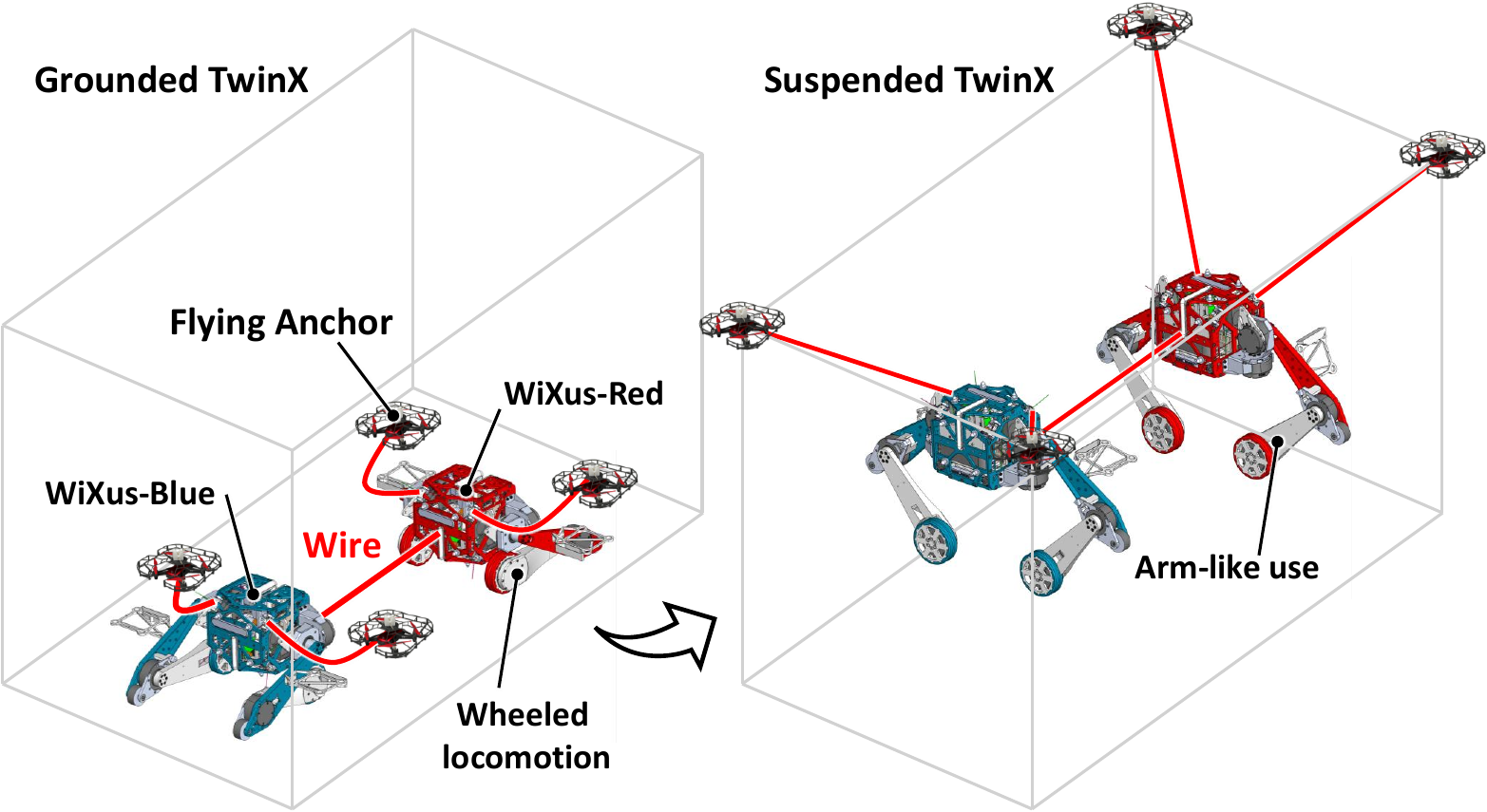}
      \vspace{-3ex}
      \caption{
        Overall hardware configuration of TwinX.
        TwinX consists of two identical robot bodies, WiXus-Blue and WiXus-Red, connected by a single inter-body wire.
        Each body is a wheeled-legged robot and can additionally connect its wires to the environment using flying anchors.
      }
      \vspace{-4ex}
      \label{fig:hardware}
    \end{center}
  \end{figure}

  The hardware configuration of TwinX is shown in \figref{fig:hardware}.
  TwinX is a robot composed of two identically configured wheeled-legged robots, WiXus\cite{inoue2026wixus},
  connected by a single inter-body wire.
  Each robot is equipped with computers and sensors required for autonomous operation,
  such as a computer, a battery, and cameras.
  Hereafter, the two bodies are referred to as WiXus-Blue and WiXus-Red according to their colors.
  The inter-body wire is wound by the wire-winding module mounted on WiXus-Blue,
  and its maximum continuous tension is approximately 400 N.
  Each body is equipped with a ground locomotion mechanism based on wheeled legs,
  and at the same time is configured so that up to four wires extending from its body can be connected to and used in the environment.
  The modules that wind these wires have a maximum continuous tension of approximately 120 N.
  In addition, wire anchoring to the environment is autonomously performed
  using flying anchors\cite{inoue2024cubix}, which combine drones and anchors.

  The left side of \figref{fig:hardware}
  shows the state in which the two bodies move on the ground while remaining connected by the inter-body wire.
  In this state, the wheeled legs of each body are used mainly for locomotion.
  In contrast, the right side of \figref{fig:hardware}
  shows the state in which each body is supported by wires connected to the environment,
  and the bodies are held in the air.
  At this time, because the wheeled legs are released from support and locomotion through ground contact,
  the legs can be used like arms to handle objects.

  In addition, by winding the inter-body wire and drawing the two WiXus robots together,
  the two bodies can dock and realize morphological reconfiguration.
  However, this study does not treat this morphological reconfiguration as the main focus.

\section{Control Method and System Architecture} \label{sec:software}
\subsection{Controller for the Multi-Floating-Link System}

  We explain the control method for multi-floating-link parallel wire actuation.
  The following discussion can be handled in the same way for cases with more than two bodies
  and for cases in which they are connected by multiple wires,
  but for simplicity, with TwinX in mind,
  we consider a system in which two WiXus bodies connected to the environment are connected by a single inter-body wire.

  First,
  we describe how to construct the Jacobian $\bm{W}$ and how to transform the wire tension $\bm{f}$ into the wrench $\bm{w}$.
  Even in multi-floating-link parallel wire actuation, the following relation holds among $\bm{W}$, $\bm{f}$, and $\bm{w}$.
  \begin{equation}
    \bm{w} = \bm{W}\bm{f}
  \end{equation}
  Here, let the number of wires by which WiXus-Blue is connected to the environment be $l$,
  and the number of wires by which WiXus-Red is connected to the environment be $m$.
  Then, the wire tension $\bm{f}$ takes the following form.
  \begin{equation}
    \bm{f} = \left[f_1^\mathrm{blue}, \dots, f_l^\mathrm{blue}, f_\mathrm{inter}, f_1^\mathrm{red}, \dots, f_m^\mathrm{red}\right]^\top \in \mathbb{R}^{l+1+m}
  \end{equation}
  Here, $f_i^\mathrm{blue}$ is the tension of the $i$-th wire by which WiXus-Blue is connected to the environment,
  $f_i^\mathrm{red}$ is the tension of the $i$-th wire by which WiXus-Red is connected to the environment,
  and $f_\mathrm{inter}$ is the tension of the inter-body wire.
  In addition, the wrench $\bm{w}$ takes the following form using
  the wrench $\bm{w}_\mathrm{blue}$ acting on WiXus-Blue and
  the wrench $\bm{w}_\mathrm{red}$ acting on WiXus-Red.
  \begin{equation}
    \bm{w} =
      \begin{bmatrix}
          \bm{w}_\mathrm{blue}\\
          \bm{w}_\mathrm{red}
      \end{bmatrix}
    \in \mathbb{R}^{12 \times 1}
  \end{equation}
  In this case, the Jacobian $\bm{W}$ can be written as follows.
  \begin{equation}
    \bm{W} = \begin{bmatrix}
      \begin{array}{ccc}
      \bm{W}_\mathrm{blue} &\bm{W}^\mathrm{blue}_\mathrm{inter} &\bm{0} \\
      \bm{0} &\bm{W}^\mathrm{red}_\mathrm{inter} &\bm{W}_\mathrm{red}
      \end{array}
      \end{bmatrix}
      \in \mathbb{R}^{12 \times (l+1+m)}
  \end{equation}
  \begin{figure}[t]
    \begin{center}
      \includegraphics[width=0.6\columnwidth]{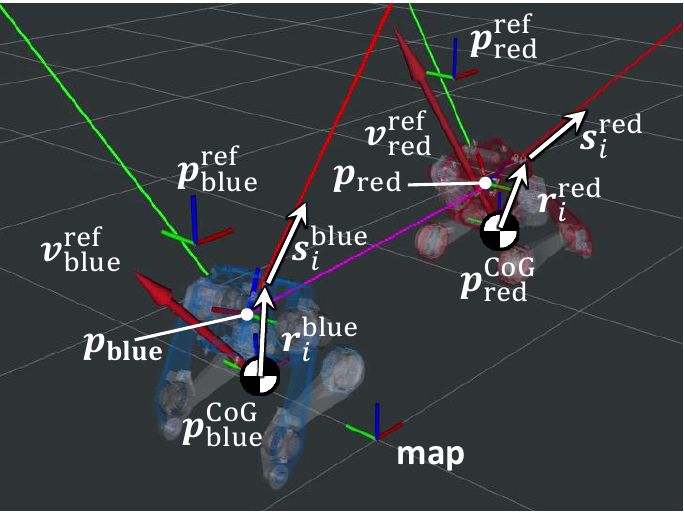}
      \vspace{-1ex}
      \caption{
        Definition of symbols used in the multi-link parallel wire-driven controller.
      }
      \vspace{-4ex}
      \label{fig:twinx-multi-parallel-character}
    \end{center}
  \end{figure}
  Here,
  $\bm{W}_\mathrm{blue}$ is the Jacobian corresponding only to the wires by which WiXus-Blue is connected to the environment.
  Based on the standard kinematic and static modeling of cable-driven parallel mechanisms\cite{Gosselin2014},
  $\bm{W}_\mathrm{blue}$ is expressed as follows using the position vectors $\bm{r}_i^\mathrm{blue}$ of the wire origins
  and the wire direction vectors $\bm{s}_i^\mathrm{blue}$ shown in \figref{fig:twinx-multi-parallel-character}.
  \begin{equation}
      \bm{W}_\mathrm{blue} =
      \left[ \begin{array}{ccc}
        \bm{s}_{1}^\mathrm{blue} & \cdots & \bm{s}_{l}^\mathrm{blue} \\
        \bm{r}_{1}^\mathrm{blue}\times\bm{s}_{1}^\mathrm{blue} & \cdots & \bm{r}_{l}^\mathrm{blue}\times\bm{s}_{l}^\mathrm{blue}
      \end{array} \right]
      \in \mathbb{R}^{6 \times l}
  \end{equation}
  $\bm{W}_\mathrm{red}$ is defined in the same way for WiXus-Red.
  \begin{equation}
      \bm{W}_\mathrm{red} =
      \left[ \begin{array}{ccc}
        \bm{s}_{1}^\mathrm{red} & \cdots & \bm{s}_{m}^\mathrm{red} \\
        \bm{r}_{1}^\mathrm{red}\times\bm{s}_{1}^\mathrm{red} & \cdots & \bm{r}_{m}^\mathrm{red}\times\bm{s}_{m}^\mathrm{red}
      \end{array} \right]
      \in \mathbb{R}^{6 \times m}
  \end{equation}
  In addition, $\bm{W}^\mathrm{blue}_\mathrm{inter}$
  is the Jacobian term for the inter-body wire in WiXus-Blue,
  expressed, in the same way as for the environment-anchoring wires,
  using the position vectors that describe the wire origin and wire direction.
  \begin{equation}
      \bm{W}^\mathrm{blue}_\mathrm{inter} =
      \begin{bmatrix}
        \bm{s}_\mathrm{inter}^\mathrm{blue} \\
        \bm{r}_\mathrm{inter}^\mathrm{blue}\times\bm{s}_\mathrm{inter}^\mathrm{blue}
      \end{bmatrix}
      \in \mathbb{R}^{6 \times 1}
  \end{equation}
  Here, $\bm{r}_\mathrm{inter}^\mathrm{blue}$ and $\bm{s}_\mathrm{inter}^\mathrm{blue}$
  are the position vectors that describe the wire origin and wire direction of the inter-body wire.
  $\bm{W}^\mathrm{red}_\mathrm{inter}$ is defined in the same way for WiXus-Red.
  Because the wire being handled is the same inter-body wire as above,
  $\bm{s}_\mathrm{inter}^\mathrm{red}$ is the opposite vector of $\bm{s}_\mathrm{inter}^\mathrm{blue}$.
  \begin{equation}
      \bm{W}^\mathrm{red}_\mathrm{inter} =
      \begin{bmatrix}
        \bm{s}_\mathrm{inter}^\mathrm{red} \\
        \bm{r}_\mathrm{inter}^\mathrm{red}\times\bm{s}_\mathrm{inter}^\mathrm{red}
      \end{bmatrix}
      \in \mathbb{R}^{6 \times 1}
  \end{equation}
  Note that all of these terms must be described in a coordinate frame shared by both robots,
  and in TwinX we use the map frame of RTAB-Map\cite{rtabmap2019}, shown in \figref{fig:twinx-multi-parallel-character}.

  \begin{figure}[t]
    \begin{center}
      \includegraphics[width=1.0\columnwidth]{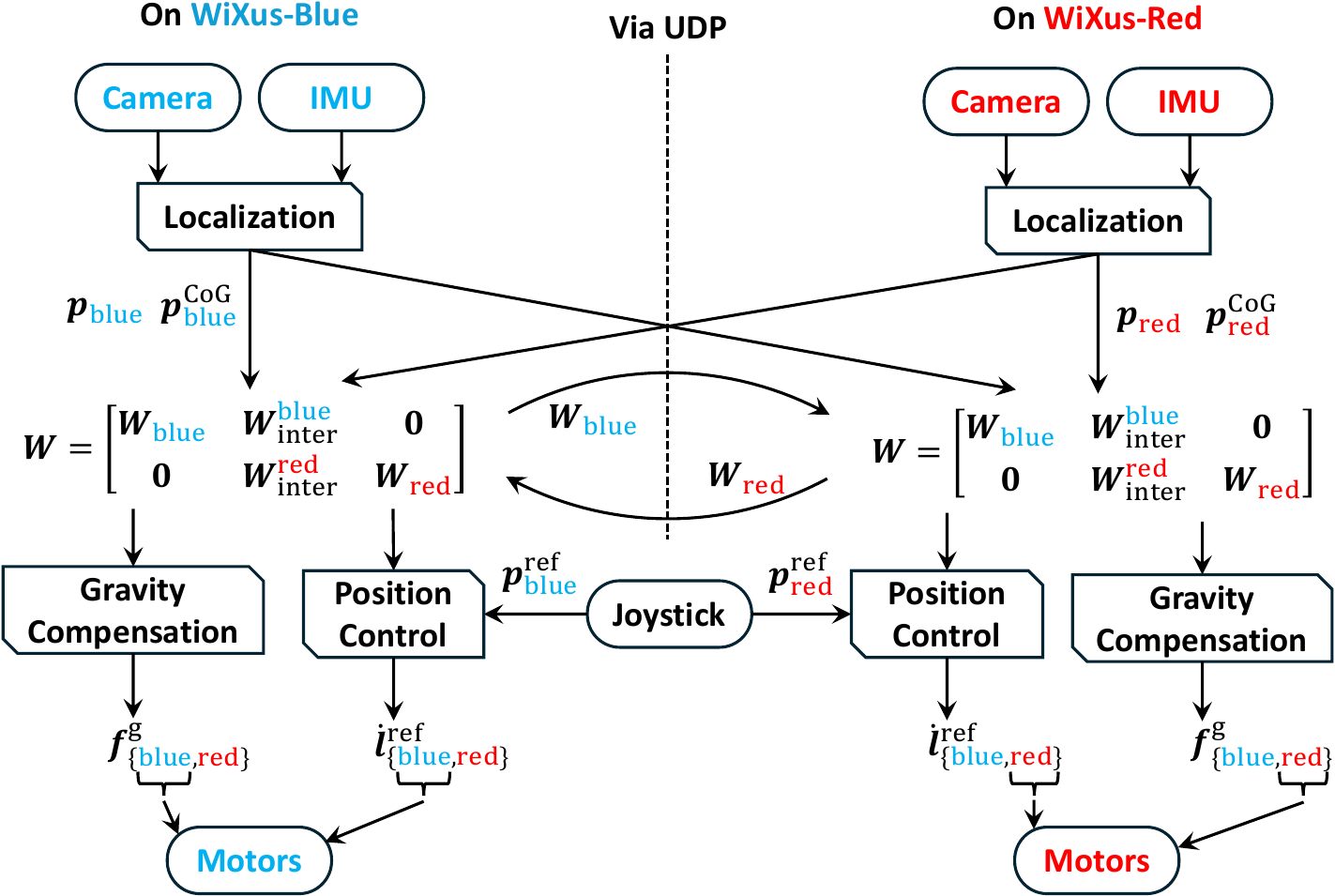}
      \vspace{-4ex}
      \caption{
        Pipeline for multi-link parallel wire-driven control.
        By exchanging localization outputs and the corresponding Jacobian components via UDP,
        both robots reconstruct and maintain the same whole-system Jacobian.
        Using this Jacobian, each robot computes gravity compensation and feedback control terms for the entire coupled system,
        and then applies only the resulting commands corresponding to its own actuators to the motors.
      }
      \vspace{-5ex}
      \label{fig:twinx-multi-parallel-flow}
    \end{center}
  \end{figure}
  Next, we explain the control that makes each WiXus follow the commanded velocity $\dot{\bm{q}}^\mathrm{ref}_\mathrm{\{blue,red\}}$.
  Using the Jacobian $\bm{W}$ constructed as above, the commanded velocity is converted into wire velocity,
  and the target wire tension $\bm{f}^\mathrm{ref}$ that tracks it
  while compensating for gravity is calculated by the following equation.
  \begin{equation}
    \bm{f}^\mathrm{ref} = \bm{f}^g + K_p\left(\dot{\bm{l}}-\bm{W}^\top\dot{\bm{q}}^\mathrm{ref}_\mathrm{TwinX}\right)
    \label{eq:twinx-cog_vel_control}
  \end{equation}
  Here, $K_p$ is the proportional gain, and $\dot{\bm{l}}$ is the vector of the wire velocities.
  The vector $\dot{\bm{q}}^\mathrm{ref}_\mathrm{TwinX}$ is obtained by stacking
  the six-dimensional commanded velocity vectors $\dot{\bm{q}}^\mathrm{ref}_\mathrm{\{blue,red\}}$ of the WiXus bodies,
  each of which consists of the translational and angular velocities, as follows.
  \begin{equation}
    \dot{\bm{q}}^\mathrm{ref}_\mathrm{TwinX} = \begin{bmatrix}\dot{\bm{q}}^\mathrm{ref}_\mathrm{blue}\\ \dot{\bm{q}}^\mathrm{ref}_\mathrm{red}\end{bmatrix}
    \in \mathbb{R}^{12 \times 1}
  \end{equation}
  By PD control that brings the center-of-mass positions of the entire WiXus bodies, $\bm{p}^\mathrm{CoG}_\mathrm{\{blue, red\}}$,
  which are computed from the robot model including the mass and center-of-gravity position of each link,
  and the current joint angles, closer to the target positions
  $\bm{p}^\mathrm{ref}_\mathrm{\{blue, red\}}$,
  the above $\dot{\bm{q}}^\mathrm{ref}_\mathrm{\{blue, red\}}$ is calculated as follows.
  Here, ${\bm{v}}_\mathrm{\{blue, red\}}$ is the translational velocity,
  ${\bm{v}}^\mathrm{ref}_\mathrm{\{blue, red\}}$ is the target translational velocity,
  and $K_P$ and $K_D$ are the proportional and derivative gains, respectively.
  \begin{equation}
    \begin{aligned}
      \dot{\bm{q}}^\mathrm{ref}_\mathrm{\{blue, red\}} &=
      \begin{bmatrix}{\bm{v}}^\mathrm{ref}_\mathrm{\{blue, red\}}\\\bm{0}_{3\times1}\end{bmatrix} \\
        &= \begin{bmatrix}K_P\left({\bm{p}}^\mathrm{ref}_\mathrm{\{blue, red\}}-{\bm{p}}^\mathrm{CoG}_\mathrm{\{blue, red\}}\right) - K_D \bm{v}_\mathrm{\{blue, red\}}\\\bm{0}_{3\times1}\end{bmatrix}
    \end{aligned}
  \end{equation}
  In addition,
  the first term of \equref{eq:twinx-cog_vel_control} is the gravity-compensation term,
  and is calculated by solving the following quadratic programming problem.
  \begin{equation}
    \begin{aligned}
      & \underset{\bm{f}^\mathrm{g}}{\text{min}}
      & & \|\bm{w}^g_\mathrm{TwinX}-{\bm{W}}{\bm{f}^\mathrm{g}}\|^2\\
      & \text{s.t.}
      & & \bm{f}^\mathrm{min} \leq \bm{f}^\mathrm{g} \leq \bm{f}^\mathrm{max} \\
    \end{aligned}
  \end{equation}
  Here,
  $\bm{f}^\mathrm{min}$ and $\bm{f}^\mathrm{max}$ are the minimum and maximum wire-tension vectors.
  $\bm{w}^g_\mathrm{TwinX}$ can be written as follows using the mass $M$ of WiXus and the gravity acceleration vector $\bm{g}$.
  \begin{equation}
    \bm{w}^g_\mathrm{TwinX} = \begin{bmatrix} (M\bm{g})^\top & \bm{0}_{1\times3} & (M\bm{g})^\top & \bm{0}_{1\times3} \end{bmatrix}^{\top}
  \end{equation}

\subsection{Execution Flow on the Real Robot}

  The processing flow of multi-floating-link parallel wire-driven control
  on the real TwinX is shown in \figref{fig:twinx-multi-parallel-flow}.
  Each robot locally executes everything from self-localization by RTAB-Map\cite{rtabmap2019}
  to the control computation described above.
  Because the whole-system Jacobian $\bm{W}$ includes components that depend on the other robot,
  the required localization and Jacobian components are exchanged via UDP.
  The UDP communication thread ran at 200 Hz, while the main control thread ran at 50 Hz.
  Each robot constructed $\bm{W}$ using the latest received values,
  and no explicit delay compensation or packet-loss recovery was implemented.
  In a round-trip latency measurement on the same Wi-Fi network used in the physical experiment,
  the round-trip time was 4.1 ms on average and 5.9 ms at maximum.
  Although the control computation produces outputs for all wires,
  each robot extracts only the components corresponding to the wires driven by itself and uses them as motor inputs.
  Because the inter-body wire is driven by WiXus-Blue, that component is applied on the WiXus-Blue side.
  In addition, the target positions are updated by joystick operation.

\section{Experiments} \label{sec:experiments}
\subsection{Simulation-Based Performance Evaluation} \label{subsec:exp_rescue}

\begin{figure}[t]
    \begin{center}
      \includegraphics[width=1.0\columnwidth]{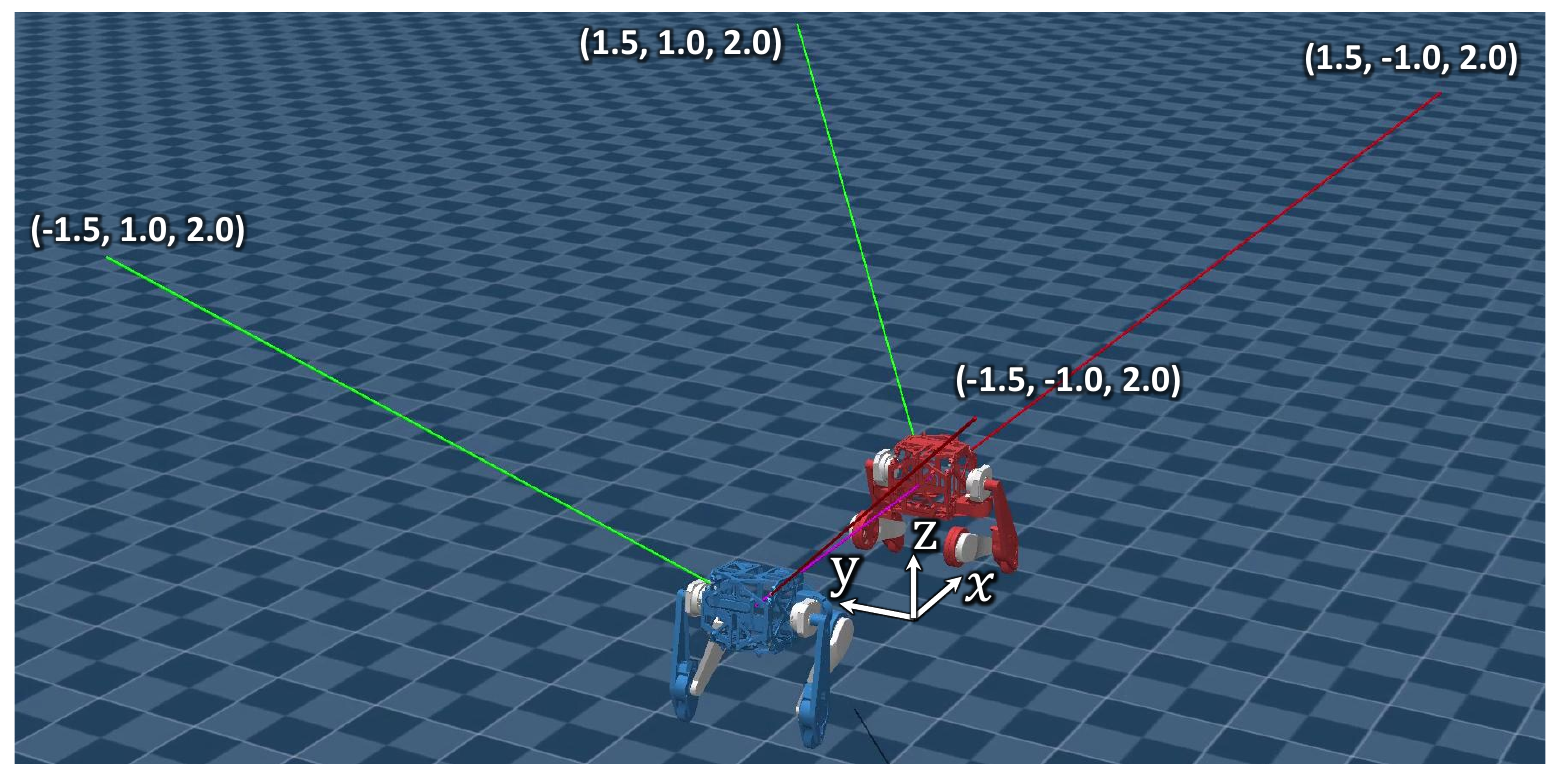}
      \vspace{-4ex}
      \caption{
        Simulation setup for the performance evaluation.
        The world coordinate frame is defined as shown in the figure.
        Four wires are attached to anchor points.
        WiXus-Blue and WiXus-Red are additionally connected by an inter-body wire.
      }
      \vspace{-4ex}
      \label{fig:exp-mujoco-pic}
    \end{center}
  \end{figure}

  To confirm that TwinX can follow the target body positions without breakdown of the controller,
  we conducted a performance evaluation in the MuJoCo simulator\cite{todorov2012mujoco}.
  The simulation setup is shown in \figref{fig:exp-mujoco-pic}.
  The coordinate frame is defined as shown in the figure,
  and four anchor points for environment anchoring,
  $(1.5, 1.0, 2.0)$, $(1.5, -1.0, 2.0)$, $(-1.5, 1.0, 2.0)$, and $(-1.5, -1.0, 2.0)$, were used.
  WiXus-Blue and WiXus-Red are additionally connected by a single inter-body wire.

  In the evaluation, inputs were given so that the target positions of each body were switched sequentially.
  The trajectories between target points were interpolated with min-jerk trajectories\cite{Flash1688}.
  In this setting,
  motions in which the same translational target was given to both bodies and the relative position was preserved
  were denoted as \textit{keep},
  whereas motions in which different targets were given to the two bodies and the relative position was changed
  were denoted as \textit{change}.
  In addition, the interval in which the reference position changed toward each target point was defined as \textit{MOVE},
  and the interval in which the reference position remained constant was defined as \textit{HOLD}.
  Note that inputs that changed the relative position in the $z$ direction were excluded from the evaluation,
  because they clearly went outside the body-position pairs that are mechanically feasible under the present wire arrangement.

\begin{figure}[t]
    \begin{center}
      \includegraphics[width=1.0\columnwidth]{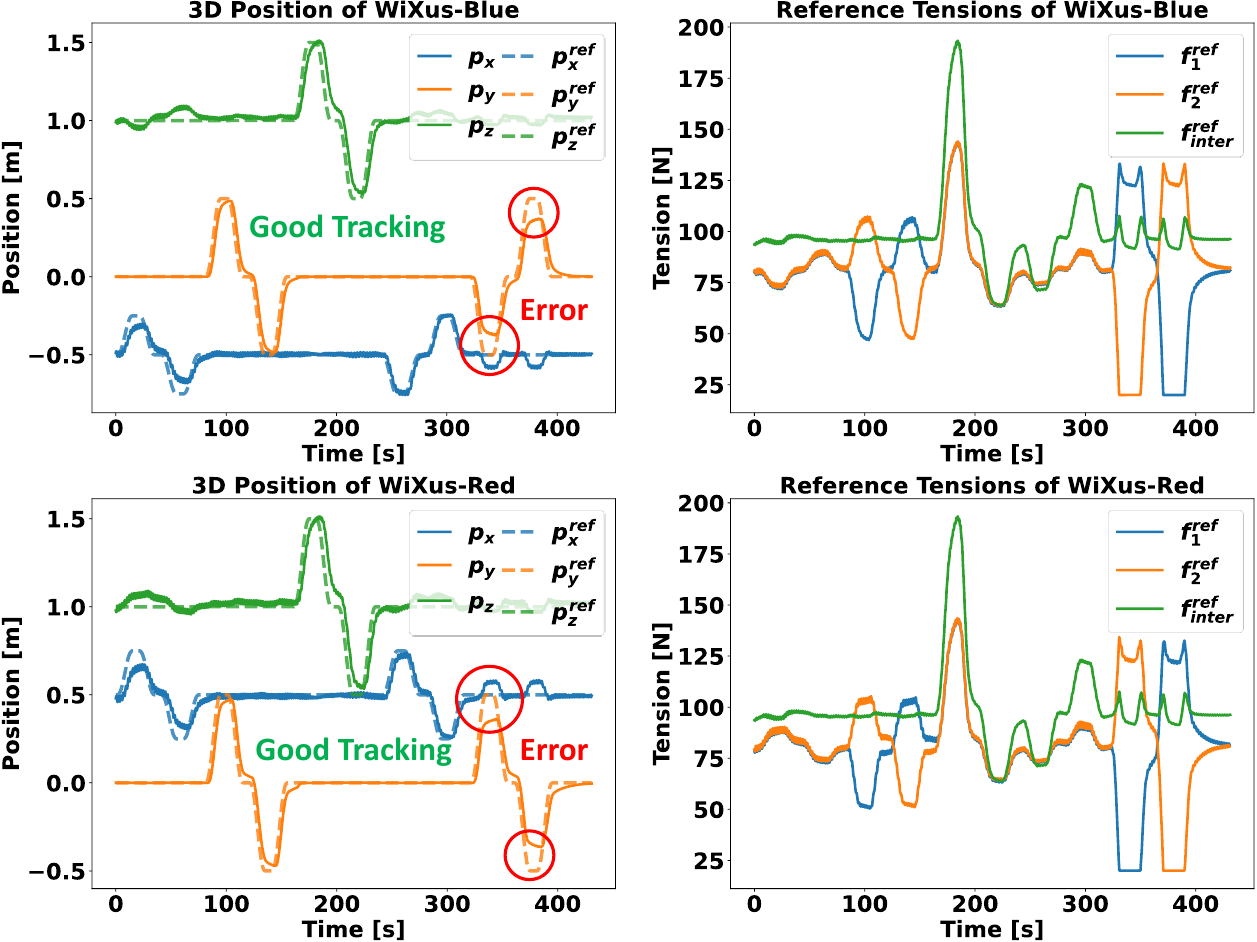}
      \vspace{-4.5ex}
      \caption{
        Tracking results in simulation.
        The left column shows the measured and reference body positions, and the right column shows the reference wire tensions.
        For each robot, $f_1^{\mathrm{ref}}$ and $f_2^{\mathrm{ref}}$ denote the tensions of the environment-anchoring wires
        connected to the $-y$ and $+y$ anchors, respectively;
        the same $f_{\mathrm{inter}}^{\mathrm{ref}}$ is shown in both plots, yielding five distinct wire tensions.
      }
      \vspace{-4.5ex}
      \label{fig:exp-mujoco-data}
    \end{center}
  \end{figure}

  The tracking results are shown in \figref{fig:exp-mujoco-data}.
  The left column shows time series of the measured and target positions of WiXus-Blue and WiXus-Red,
  and the right column shows the corresponding target tensions.
  Although each body exhibits errors in some intervals,
  it can be seen that, overall, the control tracks the target position sequence without breakdown.
  In particular, the intervals in which large errors appear correspond to intervals in which the target positions are close to,
  or outside, the feasible body-position pairs under the given wire arrangement,
  and can therefore be interpreted as resulting from the mechanical constraints of this configuration.

\begin{figure}[t]
    \begin{center}
      \includegraphics[width=1.0\columnwidth]{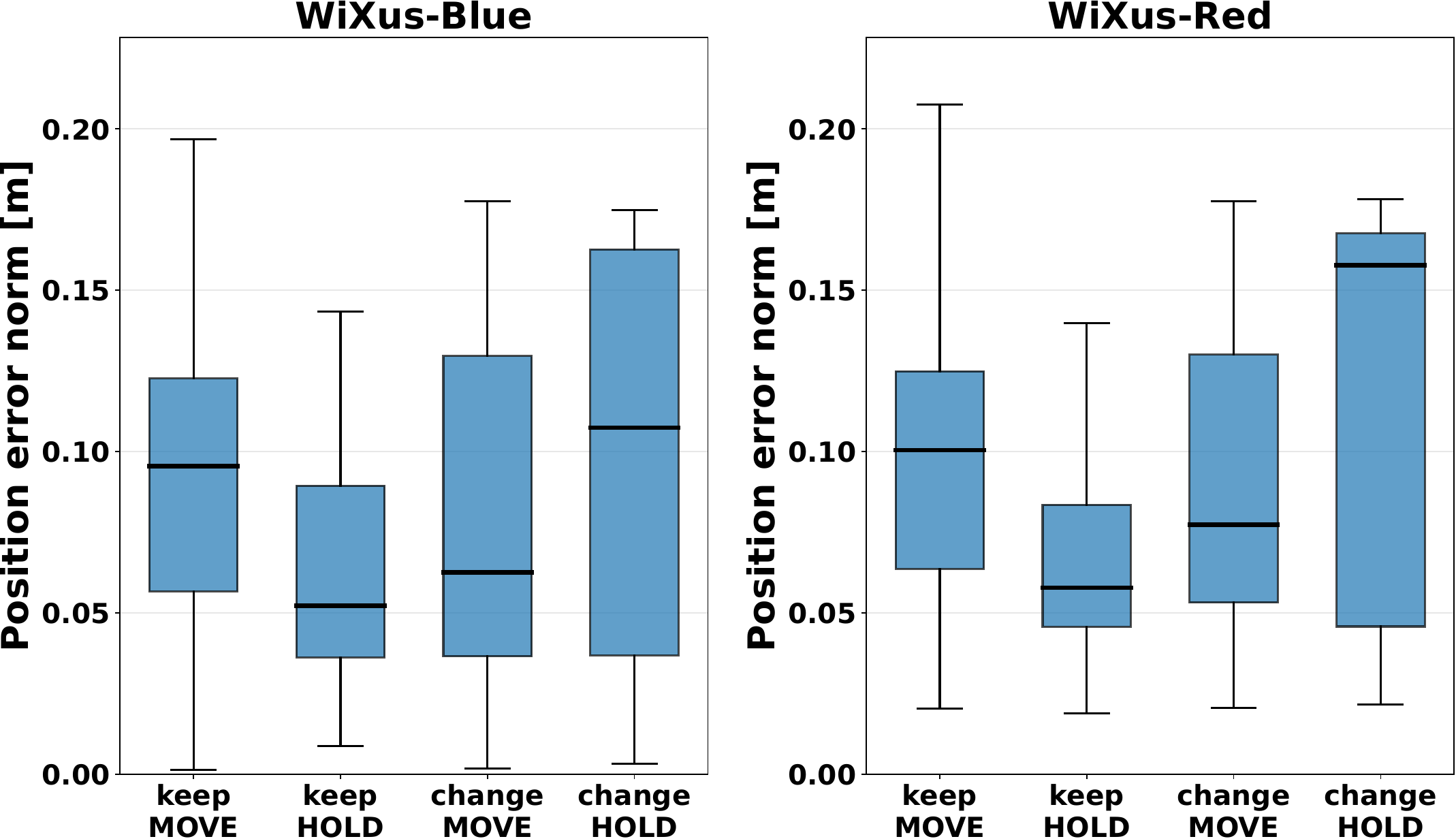}
      \vspace{-3ex}
      \caption{
        Boxplots of the position error norm for WiXus-Blue and WiXus-Red,
        grouped by whether the relative position was preserved (\textit{keep}) or changed (\textit{change}),
        and by whether the robot was moving toward a target (\textit{MOVE}) or holding a target position (\textit{HOLD}).
      }
      \vspace{-5ex}
      \label{fig:exp-mujoco-boxplot}
    \end{center}
  \end{figure}

  Next, \figref{fig:exp-mujoco-boxplot} shows boxplots of the position error norm,
  grouped by \textit{keep / change} and \textit{MOVE / HOLD}.
  For both WiXus-Blue and WiXus-Red, the errors remain roughly on the order of $10^{-1}\,\mathrm{m}$.
  This is insufficient for tasks that require precise positioning,
  but can be regarded as an applicable level of accuracy for tasks such as rough transport or towing.
  In addition, no strong tendency was observed for the error characteristics to differ greatly
  between \textit{keep / change} or between \textit{MOVE / HOLD},
  showing that the proposed controller can be applied with a similar level of stability
  both when the relative position is preserved and when it is changed.

\subsection{Real-Robot Rescue Experiment} \label{subsec:exp_tool}

  In this experiment,
  we conduct a rescue experiment as an application example of multi-floating-link parallel wire actuation using the real TwinX.

\begin{figure*}[t]
    \begin{center}
      \includegraphics[width=1.77\columnwidth]{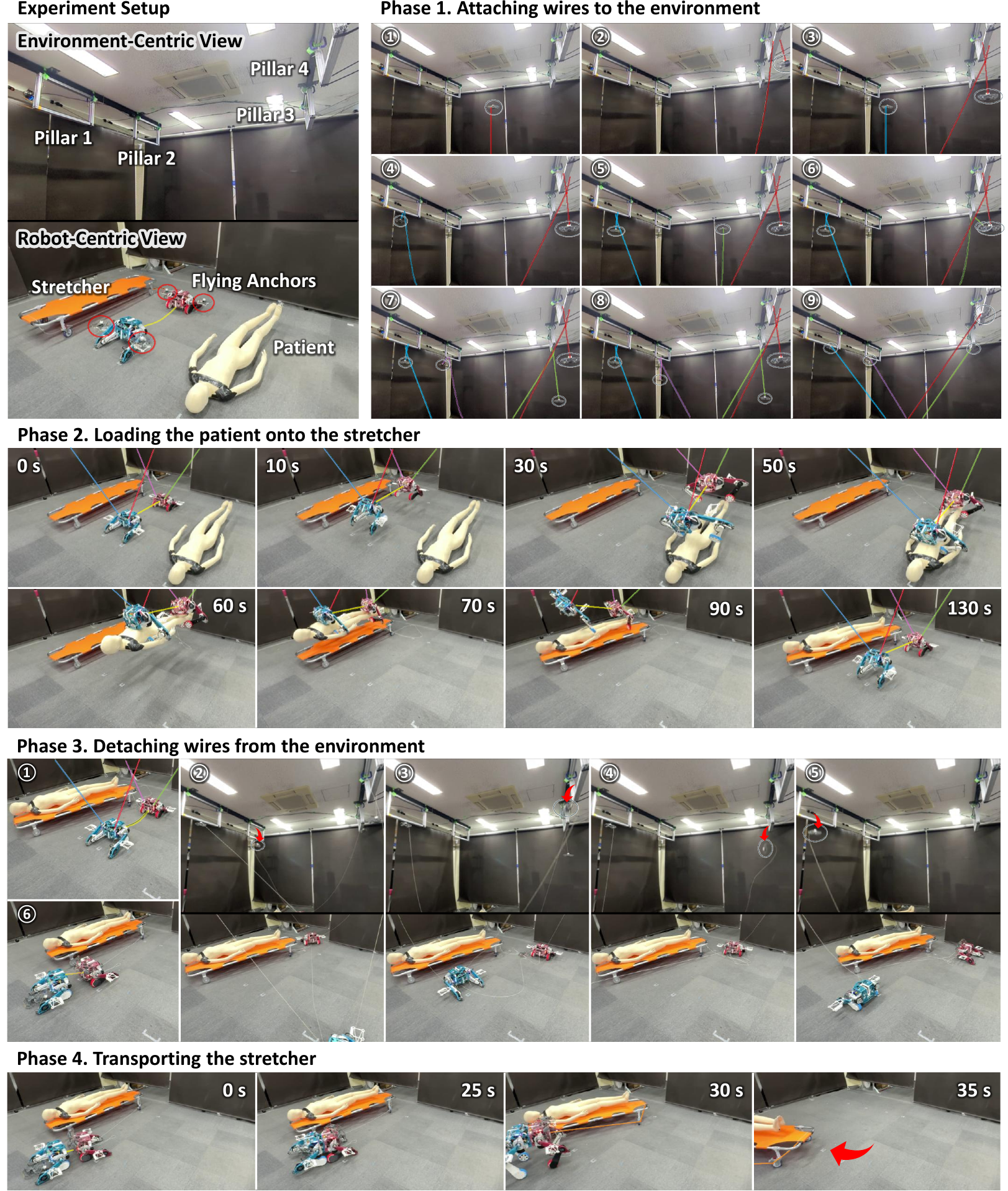}
      \vspace{-2ex}
      \caption{
        Real-robot rescue-task demonstration.
        Phase 1: four wires are attached to ceiling pillars using flying anchors.
        Phase 2: TwinX lifts the patient at the waist and legs and transfers the patient onto the stretcher by multi-floating-link parallel wire actuation.
        Phase 3: the wires are detached and recovered.
        Phase 4: TwinX reconfigures into the vehicle configuration and tows the stretcher.
      }
      \vspace{-4ex}
      \label{fig:twinx-exp-rescue-pic}
    \end{center}
  \end{figure*}

  First, we explain the scenario of the rescue experiment.
  As shown at the upper left of \figref{fig:twinx-exp-rescue-pic},
  the experimental environment includes four pillars fixed to the ceiling,
  a patient to be rescued (a mannequin with a mass of 5 kg and a height of 150 cm), and a stretcher.
  TwinX aims to place the patient on the stretcher and carry the patient out.
  The stretcher is a tool handled by TwinX, and is connected in advance by a wire deployed from WiXus-Red.
  The purpose of this experiment is not only to demonstrate the transportation of a rescue target by two robots,
  but also to demonstrate a sequence of operations in which the robots deploy environment-anchoring wires at the task site,
  construct a parallel wire-driven system including multiple floating links,
  support the target at multiple points, lift it, transfer it onto a stretcher, and tow the stretcher.
  The flying anchors are one implementation for deploying the environment-anchoring wires on site,
  and we assume an environment where anchorable objects such as columns, beams, or trees are available.
  If pre-installed anchors or manually installed anchors are used,
  the wire deployment procedure in Phase 1, described below, can be replaced.

\begin{figure}[t]
    \begin{center}
      \includegraphics[width=0.8\columnwidth]{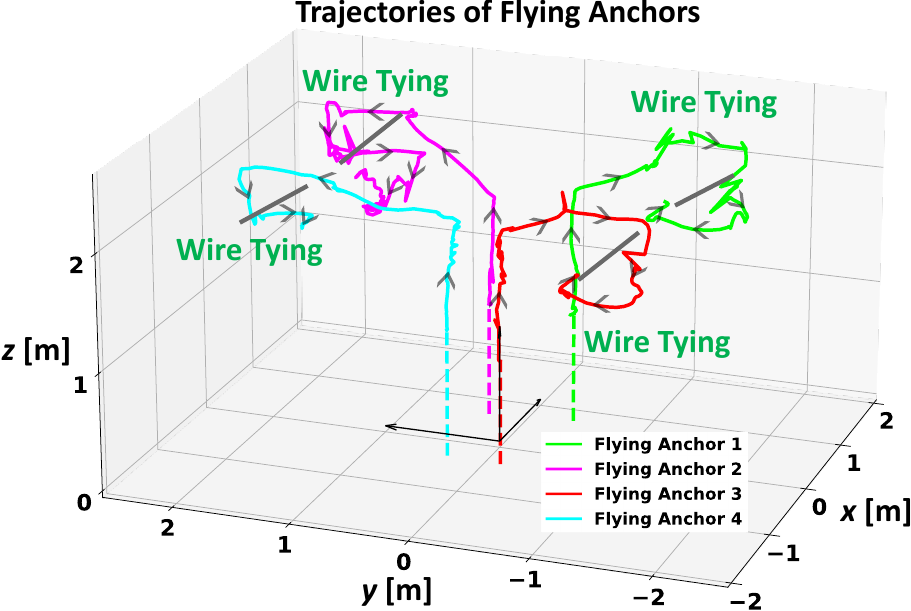}
      \vspace{-2ex}
      \caption{
        Trajectories of the four flying anchors while attaching the wires to the environment
        during the integrated rescue-task experiment.
      }
      \vspace{-2ex}
      \label{fig:twinx-exp-rescue-data-anchor}
    \end{center}
  \end{figure}

  \figref{fig:twinx-exp-rescue-pic} shows the results in Phases 1 to 4.
  In Phase 1,
  four flying anchors connected two wires from each WiXus to the four ceiling pillars
  using the RGB-D camera-based method in \cite{inoue2025anchor}.
  The trajectories of the flying anchors are shown in \figref{fig:twinx-exp-rescue-data-anchor}.
  All flying anchors were observed and controlled from WiXus-Red,
  and both WiXus robots localized themselves on a common RTAB-Map.
  The pillar positions were obtained in advance by observing the pillars with WiXus\cite{inoue2025anchor}
  and used as the environmental anchor points in $\bm{W}$;
  the actual wire-fixing positions after attachment were not re-estimated.
  Therefore, errors in the wire-fixing positions remained as modeling errors in the anchor-point positions.

\begin{figure}[t]
    \begin{center}
      \includegraphics[width=1.0\columnwidth]{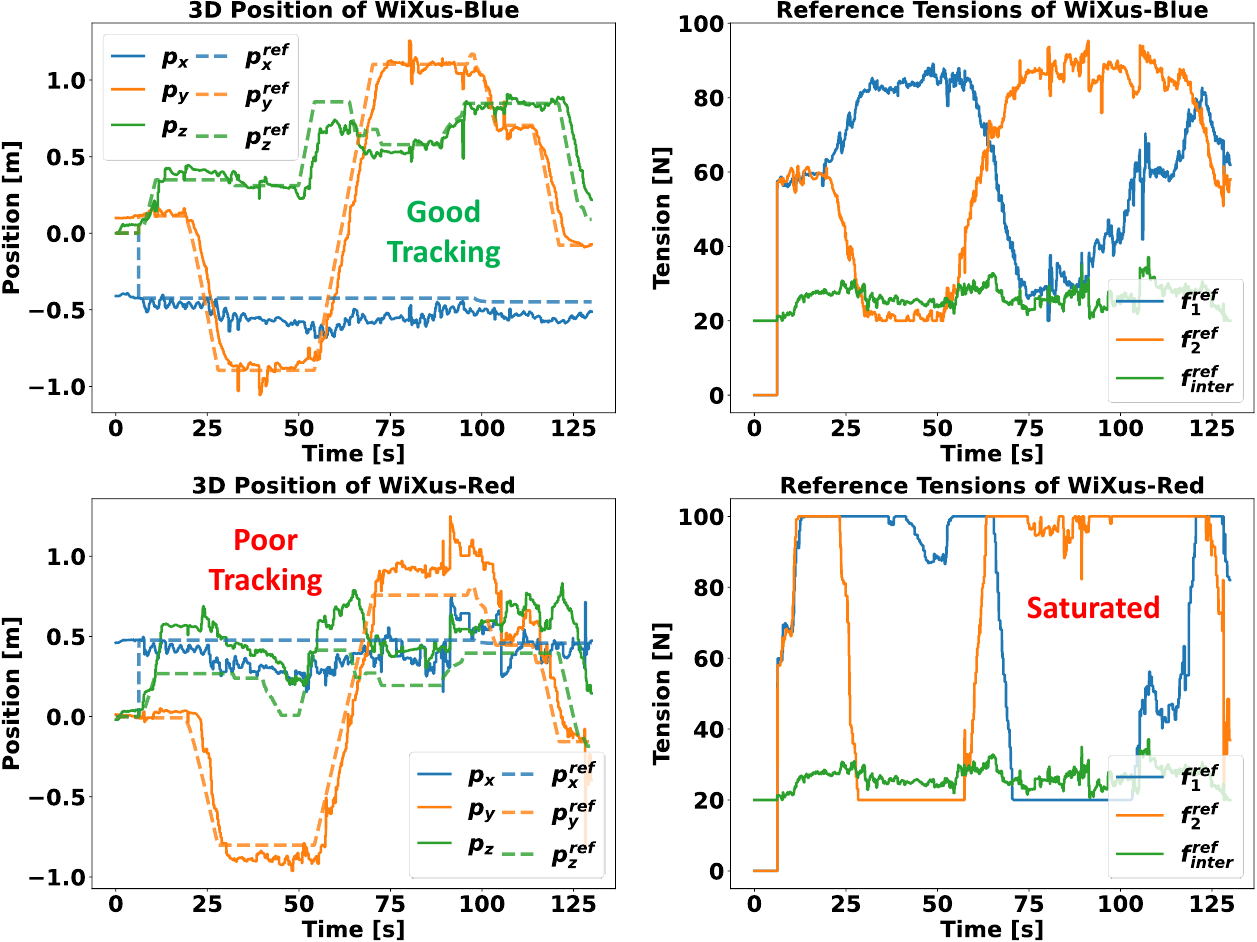}
      \vspace{-4ex}
      \caption{
        Time-series body positions and reference wire tensions during multi-link parallel wire driving in the integrated rescue-task experiment.
        For each robot, $f_1^{\mathrm{ref}}$ and $f_2^{\mathrm{ref}}$ denote the tensions of the environment-anchoring wires connected to the $-y$ and $+y$ anchors, respectively; the same $f_{\mathrm{inter}}^{\mathrm{ref}}$ is shown in both plots, yielding five distinct wire tensions.
      }
      \vspace{-5ex}
      \label{fig:twinx-exp-rescue-data-tension}
    \end{center}
  \end{figure}

  In Phase 2,
  the total of five wires, namely the four wires connected to the environment and the inter-body wire, were used,
  and the patient was transferred onto the stretcher
  by object manipulation using the legs as arms while moving in the air by multi-floating-link parallel wire actuation.
  The time-series positions and target tensions are shown in \figref{fig:twinx-exp-rescue-data-tension}.
  In \figref{fig:twinx-exp-rescue-data-tension},
  WiXus-Blue follows the target positions reasonably well, whereas WiXus-Red shows intervals in which tracking is insufficient.
  Looking at the commanded tensions,
  there are intervals on the WiXus-Red side in which the commanded values are saturated at the maximum tension of 100 N.
  This saturation is mainly considered to be caused by the payload being close to the supportable limit
  under the present wire arrangement.
  Each WiXus body weighs 9.4 kg, and in this experiment, four environment-anchoring wires were connected obliquely;
  therefore, the vertical force component available for supporting the system is smaller than the wire tension itself.
  By approximating the vertical component of each wire direction as $1/\sqrt{3}$,
  the net payload capacity after subtracting the mass of the two WiXus bodies is approximately 9.5 kg
  under the hardware tension limit of 120 N,
  and approximately 4.7 kg under the software tension limit of 100 N used in the experiment.
  This value is close to the 5 kg mannequin mass, which is consistent with the tension saturation observed in Phase 2.
  In this estimate, the required inter-body wire tension remains below its continuous tension limit.

  In Phase 3,
  the four wires connected to the environment were detached.
  Each WiXus was manually operated,
  and the wires were untied from the pillars by performing the reverse of the tying motion executed by the flying anchors.
  This detachment motion was performed by the operator while visually checking the robot bodies and wire routing,
  and no pre-programmed reverse trajectory was used.
  After the tie was undone,
  the flying anchors could be recovered together with the wires by pulling the wires back.

  In Phase 4,
  the stretcher carrying the patient was towed and carried out.
  TwinX first wound the inter-body wire to change into a vehicle configuration,
  and then towed the stretcher by winding the wire deployed from WiXus-Red.

\section{Discussion}

  First, we discuss the tracking performance observed in the simulation and real-robot experiments.
  The results suggest that large errors occur when the target positions approach or exceed the statically supportable range
  under the current wire arrangement.
  This is because TwinX is an underactuated system that handles multiple floating links with a small number of wires,
  and the required feedback wrench can reach the boundary of the feasible wrench set.
  In the real robot, oscillation during suspension, cable elongation, and mechanical friction can further
  affect tension generation, while errors in the environmental anchor-point positions remain as modeling errors
  in the structure matrix $\bm{W}$.
  Although these factors were not individually identified,
  avoiding states near the wrench limits is expected to reduce tracking error.
  In this paper, target positions were connected by min-jerk trajectories \cite{Flash1688},
  but the current path generation does not check wrench feasibility before execution.
  Thus, operator inputs or sequential targets may command states that are difficult to realize
  under the current wire arrangement.
  To avoid this, trajectory optimization considering feasible wrenches and feasible body-position pairs will be necessary.
  This direction is consistent with motion planning for underactuated CDPRs under tension constraints\cite{Oh2006,Gouttefarde2011,Ida2019};
  however, extending the driven object from a single floating link to multiple floating links makes the problem
  higher-dimensional because combinations of body configurations must be considered.
  In addition, some tasks may require optimizing anchor placement, namely wire arrangement optimization,
  because good tracking may not be achievable under an inappropriate wire arrangement regardless of the path.

  Next, we describe the significance of the rescue experiment and the remaining issues.
  The important point of this experiment is that it showed multi-floating-link parallel wire actuation
  as an integrated robotic process, rather than only as an isolated wire-driven motion.
  In particular, the experiment included both the on-site deployment of the wire-driven system and
  its use for target support and transport within a single system.
  Whereas many conventional CDPRs have been driven while relying on a predesigned fixed frame,
  this study shows that the system can be executed at the work site starting from the stage of connecting the wires.
  This reframes parallel wire actuation not as a mechanism inside a prearranged facility,
  but as a robotic system that can be brought to the required location and deployed and used on site.
  In particular, when the wires are connected to the environment by flying anchors,
  the number of wires that can be handled simultaneously in practice is likely to be limited to only a few at most.
  In general, more wires are required to sufficiently control a single floating link in three-dimensional space.
  Considering this, it is highly significant that this study constructed a multi-link system including two floating links
  and completed an actual rescue task with the realistic number of four environment-anchoring wires.
  That is, this study shows that, even under a limited and realistic number of wires,
  introducing an inter-body wire makes it possible to realize multi-floating-link parallel wire actuation
  and achieve multipoint support and transport of the target.

  On the other hand, real-robot operation also suggests that self-localization can limit the overall system performance.
  In this paper, the two WiXus robots as floating links were self-localized by RTAB-Map\cite{rtabmap2019},
  but the estimation tended to become unstable because of three-dimensional motion such as oscillation in the air.
  In principle, information such as wire lengths, known anchor points, and wire directions
  imposes geometric constraints on the system position and orientation,
  and by using these as a prior distribution or as observation constraints,
  it may be possible to realize more stable self-localization.
  In the future, rather than relying only on visual SLAM,
  it will be necessary to develop state estimation that integrates wire lengths, tensions, and known environmental structures,
  thereby increasing the robustness of self-localization during suspension.
  Such stabilization of estimation will be important
  not only for improving the reliability of the indoor experiments shown in this study,
  but also for extending the system to outdoor environments that involve larger disturbances and environmental uncertainty.
  Another implementation limitation is the communication between the two robots.
  Because the present experiments involved relatively slow motions,
  we did not observe noticeable effects of UDP communication delay during the experiments.
  However, a systematic evaluation of robustness against packet loss and communication delay remains future work.

\section{Conclusion} \label{sec:conclusion}

  In this study, we proposed TwinX as a multi-floating-link parallel wire-driven system with an inter-body wire,
  and extended CDPRs, which have conventionally focused mainly on a single floating link,
  to a framework in which multiple floating links cooperatively operate.
  Specifically, by connecting two wheeled-legged robots with a single inter-body wire,
  and using a configuration in which each body connects wires to the environment,
  we constructed a system in which multiple floating links are cooperatively driven by parallel wire actuation
  within the workspace.

  In the performance evaluation in MuJoCo,
  we confirmed that the proposed controller can realize tracking of the target position sequence without breakdown of control.
  Furthermore, in real-robot experiments, using four environment-anchoring wires and one inter-body wire,
  we continuously realized, within a single system,
  a series of rescue tasks in which the patient was supported and lifted at two points, transferred onto a stretcher,
  and then the wires were detached from the environment and towing transport was carried out.
  This shows that, by introducing an inter-body wire,
  multi-floating-link parallel wire actuation can be realized,
  and CDPRs can be extended to the transport and manipulation of targets that require multipoint support.

  Because the proposed system handles multiple floating links with a small number of wires,
  tracking performance degraded when the target positions or required wrench exceeded
  the feasible range of the current wire arrangement.
  Self-localization instability also limited the real-robot performance.
  Future work will address these issues through trajectory optimization considering feasible wrenches and body-position pairs,
  wire arrangement optimization, and state estimation that integrates wire lengths, tensions, and known environmental structures.
  The proposed framework can be extended to systems with more bodies, where wire arrangement design,
  interference avoidance, path planning, and state estimation become increasingly important
  for multipoint support of larger objects and more complex cooperative tasks.

{
  \bibliographystyle{IEEEtran}
  \bibliography{bib}
}

\end{document}